\documentclass[11pt]{article}

\usepackage[final]{acl}
\usepackage{times}
\usepackage{latexsym}

\usepackage{microtype}
\usepackage{inconsolata}

\usepackage{amsmath}
\usepackage{graphicx}   % 图片
\usepackage[table]{xcolor} % 颜色（含表格颜色支持）
\usepackage{booktabs}   % 表格线
\usepackage{tabularx}   % 自适应列宽表
\usepackage{array}      % 自定义列类型
\usepackage{multirow}   % 表格多行单元
\usepackage{enumitem}   % 列表格式
\usepackage{pifont}     % \ding 符号（你的对勾/叉需要）
\usepackage{amsfonts}
\usepackage{listings}   % 代码块
\usepackage{newfloat}   % 自定义浮动体（如 Listing）
\usepackage{algorithm}  % 伪代码环境
\usepackage{algorithmicx}
\usepackage{algpseudocode}

\usepackage{placeins}
\usepackage{float}

\usepackage[table,xcdraw]{xcolor} % 提供颜色支持
\definecolor{headergray}{gray}{0.92}
\definecolor{humanblue}{rgb}{0.96, 0.97, 1.0}

\usepackage[most]{tcolorbox}

\tcbset{
  colback=gray!3,
  colframe=gray!35,
  boxrule=0.4pt,
  arc=1pt,
  outer arc=1pt,
  left=6pt,right=6pt,top=4pt,bottom=4pt,
  before skip=6pt,after skip=6pt
}

\newtcolorbox{promptbox}[1][]{
  breakable,
  enhanced,
  fonttitle=\bfseries,
  colback=gray!3,
  colframe=gray!40,
  #1
}

\newcommand{\ResetTable}[1]{\resizebox{\linewidth}{!}{#1}}

\newcolumntype{L}{>{\raggedright\arraybackslash}p{2.5cm}}

\lstdefinelanguage{json}{
  morestring=[b]",
  showstringspaces=false,
  breaklines=true,
  alsoletter={0123456789},
  sensitive=true
}

\floatstyle{ruled}
\newfloat{listing}{tb}{lst}{}
\floatname{listing}{Listing}

\newcommand{\Func}[1]{\textsc{#1}}
\newcommand \footnoteONLYtext[1]
{
	\let \mybackup \thefootnote
	\let \thefootnote \relax
	\footnotetext{#1}
	\let \thefootnote \mybackup
	\let \mybackup \imareallyundefinedcommand
}

\title{Navigating Large-Scale Document Collections: MuDABench for Multi-Document Analytical QA}
\author{
  Zhanli Li\textsuperscript{1,3} \quad
  Yixuan Cao\textsuperscript{1,2}\thanks{Corresponding Author: Yixuan Cao.} \quad
  Lvzhou Luo\textsuperscript{1,2} \quad
  Ping Luo\textsuperscript{1,2} \\
  \textsuperscript{1}State Key Laboratory of AI Safety, Institute of Computing Technology,\\ Chinese Academy of Sciences (CAS), Beijing 100190, China \\
  \textsuperscript{2}University of Chinese Academy of Sciences, Beijing 100049, China \\
  \textsuperscript{3}Wenlan School of Business, Zhongnan University of Economics and Law, Wuhan 430073, China \\
  \texttt{lizhanli@stu.zuel.edu.cn} \quad
  \texttt{\{caoyixuan, luolvzhou23s, luop\}@ict.ac.cn}
}

\begin{document}
\maketitle

\begin{abstract} 
This paper introduces the task of analytical question answering over large, semi-structured document collections. We present MuDABench, a benchmark for multi-document analytical QA, where questions require extracting and synthesizing information across numerous documents to perform quantitative analysis. Unlike existing multi-document QA benchmarks that typically require information from only a few documents with limited cross-document reasoning, MuDABench demands extensive inter-document analysis and aggregation.
Constructed via distant supervision by leveraging document-level metadata and annotated financial databases, MuDABench comprises over 80,000 pages and 332 analytical QA instances. We also propose an evaluation protocol that measures final answer accuracy and uses intermediate-fact coverage as an auxiliary diagnostic signal for the reasoning process. Experiments reveal that standard RAG systems, which treat all documents as a flat retrieval pool, perform poorly. To address these limitations, we propose a multi-agent workflow that orchestrates planning, extraction, and code generation modules. While this approach substantially improves both process and outcome metrics, a significant gap remains compared to human expert performance. Our analysis identifies two primary bottlenecks: single-document information extraction accuracy and insufficient domain-specific knowledge in current systems. MuDABench is available at \url{https://github.com/Zhanli-Li/MuDABench}.
\end{abstract}

\footnoteONLYtext{This work was done during Zhanli Li's internship at the Institute of Computing Technology, Chinese Academy of Sciences.}


\section{Introduction}

Large language models (LLMs) combined with retrieval-augmented generation (RAG) are now the dominant paradigm for question answering over unstructured content such as the web, enterprise knowledge bases, and document repositories~\citep{gao2023retrieval}. In most settings, these systems treat documents as loosely-related snippets: the goal is to retrieve a small set of passages that fit into a single context window and then answer the query in one or a few model calls. Wikipedia-style multi-hop datasets such as HotpotQA and its successors~\cite{yang2018hotpotqa,ho2020constructing,trivedi2022musique,zhu2024fanoutqa,levy2025more} instantiate this view, and recent work on long-context benchmarks extends it to longer inputs without changing the underlying interaction pattern.
\begin{figure}[t]
    \centering
    \includegraphics[width=1\linewidth]{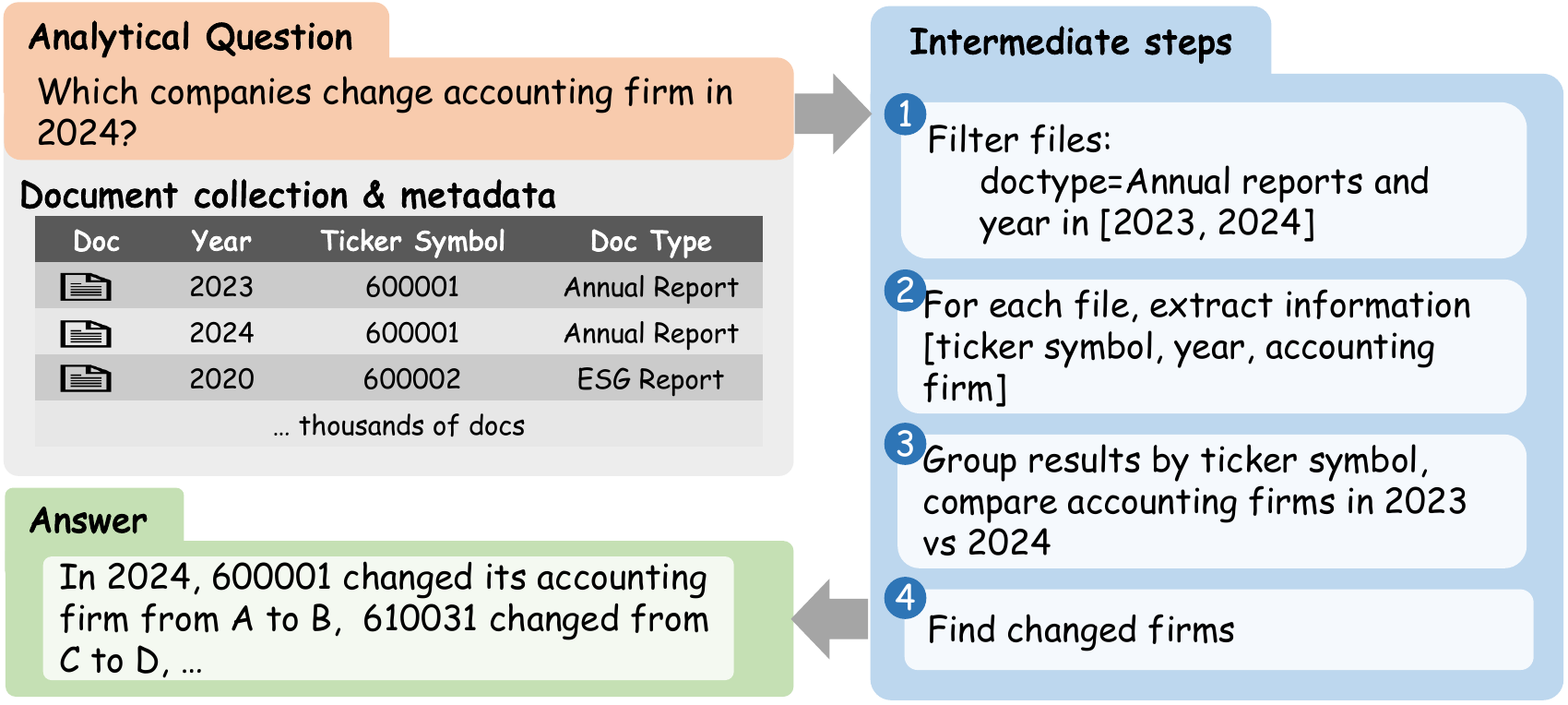}
    \caption{An example of multi-doc analytical QA. The collection of documents behind a question is organized into a semi-structured database through metadata, and answering the question involves first identifying which documents are useful and then targeting the information extraction for the final aggregated answer.}
    \label{fig:example}
\end{figure}

\begin{table*}[t]

\resizebox{\textwidth}{!}{%
\begin{tabular}{lccccccc}
\toprule
\textbf{Dataset} & \textbf{Doc / Q} & \textbf{Pages / Doc} & \textbf{Multihop} &\textbf{Metadata}& \textbf{Multilingual}&\textbf{Realistic}&  \textbf{Publicly released}\\
\midrule
 HotpotQA \citep{yang2018hotpotqa}& $\sim$ 3 webs& -& \textcolor{green}{{\Large \ding{51}}}& \textcolor{red}{{\Large \ding{55}}}& \textcolor{red}{{\Large \ding{55}}}& \textcolor{green}{{\Large \ding{51}}}&\textcolor{green}{{\Large \ding{51}}}\\
2WikiMultiHopQA \citep{ho2020constructing} & $\sim$ 3 webs& - & \textcolor{green}{{\Large \ding{51}}} &\textcolor{red}{{\Large \ding{55}}}& \textcolor{red}{{\Large \ding{55}}}&\textcolor{green}{{\Large \ding{51}}}&  \textcolor{green}{{\Large \ding{51}}}\\
MuSiQue \citep{trivedi2022musique} & $\sim$ 2-4 webs& - & \textcolor{green}{{\Large \ding{51}}} &\textcolor{red}{{\Large \ding{55}}}& \textcolor{red}{{\Large \ding{55}}}&\textcolor{red}{{\Large \ding{55}}}&  \textcolor{green}{{\Large \ding{51}}}\\
FanOutQA \citep{zhu2024fanoutqa} & $\sim$ 5-7 webs& - & \textcolor{green}{{\Large \ding{51}}} &\textcolor{red}{{\Large \ding{55}}}& \textcolor{red}{{\Large \ding{55}}}&\textcolor{green}{{\Large \ding{51}}}&  \textcolor{green}{{\Large \ding{51}}}\\
MoreDocsSameLen \citep{levy2025more}& $\sim$  2 webs& - & \textcolor{green}{{\Large \ding{51}}} &\textcolor{red}{{\Large \ding{55}}}& \textcolor{red}{{\Large \ding{55}}}&\textcolor{red}{{\Large \ding{55}}}& \textcolor{green}{{\Large \ding{51}}}\\
\hline
Financebench \citep{islam2023financebench} & 1 pdf & $\sim$ 10-200 pages & \textcolor{red}{{\Large \ding{55}}} &\textcolor{green}{{\Large \ding{51}}}& \textcolor{red}{{\Large \ding{55}}}&\textcolor{green}{{\Large \ding{51}}}&  \textcolor{green}{{\Large \ding{51}}}\\
Aryn \citep{anderson2024design} & - & 4-7 pages & \textcolor{green}{{\Large \ding{51}}} &\textcolor{red}{{\Large \ding{55}}}& \textcolor{red}{{\Large \ding{55}}}&\textcolor{green}{{\Large \ding{51}}}&  \textcolor{red}{{\Large \ding{55}}}\\
DocETL \citep{shankar2024docetl} & - & - & \textcolor{green}{{\Large \ding{51}}} &\textcolor{red}{{\Large \ding{55}}}& \textcolor{red}{{\Large \ding{55}}}&\textcolor{green}{{\Large \ding{51}}}&  \textcolor{red}{{\Large \ding{55}}}\\
 LongBench \citep{bai-etal-2024-longbench}& -& -& \textcolor{green}{{\Large \ding{51}}}& \textcolor{red}{{\Large \ding{55}}}& \textcolor{green}{{\Large \ding{51}}}& \textcolor{green}{{\Large \ding{51}}}&\textcolor{green}{{\Large \ding{51}}}\\
RULER \citep{hsieh2024ruler}& - & - & \textcolor{green}{{\Large \ding{51}}}& \textcolor{red}{{\Large \ding{55}}}& \textcolor{red}{{\Large \ding{55}}}& \textcolor{red}{{\Large \ding{55}}}& \textcolor{green}{{\Large \ding{51}}}\\
Loong \citep{wang2024leave}& $\sim$ 11 pdf & $\sim$ 30 pages & \textcolor{green}{{\Large \ding{51}}}&\textcolor{red}{{\Large \ding{55}}}& \textcolor{green}{{\Large \ding{51}}}&\textcolor{green}{{\Large \ding{51}}}& \textcolor{green}{{\Large \ding{51}}}\\
 LongDocURL \citep{deng2024longdocurl}& 1 pdf& 85.6 pages& \textcolor{green}{{\Large \ding{51}}}& \textcolor{red}{{\Large \ding{55}}}& \textcolor{red}{{\Large \ding{55}}}& \textcolor{red}{{\Large \ding{55}}}&\textcolor{green}{{\Large \ding{51}}}\\
 M3DocVQA
\citep{cho2025m3docvqa}& $\sim$1.4 pdf& $\sim$12 pages
& \textcolor{green}{{\Large \ding{51}}}& \textcolor{green}{{\Large \ding{51}}}& \textcolor{red}{{\Large \ding{55}}}& \textcolor{green}{{\Large \ding{51}}} & \textcolor{green}{{\Large \ding{51}}}\\
 FinAgentBench \citep{choi2025finagentbench}& 1 pdf & $\sim$ 100 pages & \textcolor{red}{{\Large \ding{55}}} & \textcolor{green}{{\Large \ding{51}}} & \textcolor{red}{{\Large \ding{55}}}&\textcolor{green}{{\Large \ding{51}}} & \textcolor{red}{{\Large \ding{55}}}\\
\hline
\rowcolor{headergray} % 表头背景色
\textbf{MuDABench (Ours)}& 14.8 pdf & 149.7 pages &\textcolor{green}{{\Large \ding{51}}} &\textcolor{green}{{\Large \ding{51}}}& \textcolor{green}{{\Large \ding{51}}}&\textcolor{green}{{\Large \ding{51}}}&  \textcolor{green}{{\Large \ding{51}}}\\
\bottomrule
\end{tabular}
}
\caption{Benchmark Comparison. Compared to Wikipedia-type benchmarks and benchmarks with long contexts, our benchmark has advantages in the number and size of documents as well as document structuring.}
\label{tab:dataset-comparison}
\end{table*}

However, another class of real-world document QA applications, namely, analytical QA over multi-document collections, has received limited research attention. Here, a document collection behaves like a semi-structured database: documents are complementary along dimensions such as entity, year, or document type, and answering a question requires aggregating information across dozens of filings. For example, financial regulators analyze annual reports, ESG disclosures, and corporate announcements of listed companies to detect abnormal changes in accounting firms or risk indicators; researchers survey hundreds of papers to construct performance tables over datasets and tasks; and public-sector agencies aggregate heterogeneous reports to audit policy outcomes. In these settings, missing one relevant document or misinterpreting one table can invalidate the final conclusion.

Figure~\ref{fig:example} illustrates an example of analytical QA that is of critical concern to financial regulators. The underlying data consists of annual reports from multiple companies over several years, and the question asks which companies changed their accounting firms in 2024, as this may signal significant financial changes. To answer this question, the required steps include: filtering all company annual reports from 2023 and 2024, extracting information tuples (year, company, accounting firm) from each report, then aggregating the information into records of the form (company, 2023 accounting firm, 2024 accounting firm, whether changed), and finally outputting the list of companies that made changes. Regulators can then focus on examining the financial status of these companies to detect problems early.

The key challenge of this problem is the large number of documents involved in the analysis. More specifically, not only is the document collection large, but the number of documents requiring actual data extraction is also substantial, potentially thousands of documents, which stands in stark contrast to datasets like HotPotQA~\cite{yang2018hotpotqa}. Therefore, traditional approaches that directly perform retrieval over all documents or rely on long-context methods both fail, necessitating further research tailored to this problem.

Current benchmarks do not cover this research problem. Wikipedia-based multi-hop datasets capture compositional reasoning but operate over short, homogeneous pages and small numbers of documents per question. Long-context benchmarks such as LongBench, RULER, and LongDocURL probe context-length limits, but they typically assume that all relevant content fits into a single context window. While recent efforts like M3DocVQA~\citep{cho2025m3docvqa} extend multimodal understanding to multiple documents, they operate on relatively small scales ($\sim$12 pages) compared to real-world repositories. In the financial domain, FinanceBench evaluates single-document QA~\citep{islam2023financebench}, and FinAgentBench~\citep{choi2025finagentbench} introduces ``agentic retrieval'' to precisely locate document types and chunks. However, these benchmarks focus on \emph{retrieval precision} rather than the downstream \emph{aggregation and analysis} of content from massive collections. System papers such as Aryn and DocETL propose multi-step workflows but do not release large-scale public benchmarks~\citep{anderson2024design,shankar2024docetl}. Table~\ref{tab:dataset-comparison} summarizes these trends.

This paper introduces \textbf{MuDABench}, a benchmark for \emph{multi-document analytical QA} over large collections of financial filings. Built from annual reports, ESG reports, and corporate announcements of Chinese and U.S. listed companies, MuDABench spans over 80{,}000 pages, 332 questions. For each question, we construct a document set with 15 documents on average. 
This quantity is sufficiently large such that the combined length exceeds the context window of current long-context LLMs, yet remains manageable to control the cost of LLM API calls during evaluation. We provide metadata information for these documents and annotate an intermediate data point set that captures the essential per-document facts required to answer the question.

For evaluation, we primarily focus on \emph{final} answer correctness, while also introducing a process-oriented diagnostic metric based on intermediate results. Enabled by the intermediate data point set in our dataset, this auxiliary signal is evaluated with task-specific LLM-as-judge protocols, including double-check fact-coverage estimation for standard RAG and cell-wise evaluation for document-grounded workflows.

We propose a metadata-aware multi-agent workflow that plans sub-queries, performs single-document extraction, normalizes answers into flat JSON, and aggregates them with generated analysis code. Experimental results show that ordinary RAG frameworks achieve low accuracy even with large retrieval budgets, and our workflow substantially improves final-answer accuracy while also yielding more complete intermediate extraction patterns in many cases. But all methods remain significantly below human performance. We conduct a detailed analysis and identify the key challenges of this task, including the requirement that a large number of single-document information extractions must all be correct (resulting in low overall success rates), as well as the insufficient domain-specific knowledge required for effective planning.

\section{Related Work}

There are a number of works for QA on documents, including QA on pages, images, and tables within a document, QA on a single document, and QA on multiple documents. We describe each of these works below.

\textbf{QA on Document Elements} Complex document elements, such as tables and images, pose distinct challenges for LLMs owing to their structured and visual characteristics. In the realm of document image QA, \citet{kahou2017figureqa} pioneered the FigureQA dataset, which comprises synthetic scientific-style figures, including line plots, dot-line plots, vertical and horizontal bar graphs, and pie charts. Complementing this, \citet{mathew2021docvqa} introduced DocVQA, a dataset encompassing over 12,000 document images paired with questions. Recent advancements in optical character recognition (OCR) and multimodal LLMs have facilitated effective performance by open-source models on these tasks. In the realm of Table QA, \citet{pang2024uncovering} developed the TabIS benchmark, employing single-choice questions to assess LLMs, while \citet{wu2025tablebench} created TableBench, a comprehensive dataset sourced from industry. These investigations underscore a performance disparity, with open-source models trailing behind proprietary counterparts, such as the GPT, which exhibit near-human performance in table-based reasoning.

\textbf{QA on Single Document}  Single-document QA involves a user specifying a document and using its information to answer questions. The development of long-context models and RAG systems has led to significant improvements in single-document QA results. This is particularly evident in specialized domains and multimodal contexts. For instance, in the financial domain, which requires specialized knowledge, the correctness rate of FinanceBench \citep{islam2023financebench} has increased from \citet{mafin_financebench} to 98.7\%. But there are still challenges here, in single-document multimodal QA, \citet{deng2024longdocurl} introduced LongDocURL, highlighting the challenges that document layout poses for LLMs.

\textbf{QA on Multi-Document}  Multi-document QA broadly involves utilizing both the web and specific document repositories as data sources. This task presents heightened complexity, necessitating that LLMs synthesize information and reason across disparate documents. Existing multi-document benchmarks, often primarily sourcing data from Wikipedia, frequently emphasize multi-hop reasoning problems involving multiple entities. These benchmarks highlight persistent deficiencies in LLMs' capabilities for robust multi-hop reasoning \citep{ho2020constructing, trivedi2022musique, zhu2024fanoutqa, levy2025more}. In the financial domain, FinAgentBench~\citep{choi2025finagentbench} targets the retrieval stage, evaluating whether agents can identify the correct document types and passages. Similarly, M3DocVQA~\citep{cho2025m3docvqa} addresses the challenge of visual reasoning across multiple documents. 

However, a common characteristic of much prior work is its treatment of multiple documents primarily as data sources from which relevant snippets are retrieved and aggregated into a single context for an LLM. Studies introducing benchmarks like Loong \citep{wang2024leave} and RULER \citep{hsieh2024ruler} reveal significant limitations in current long-context LLMs specifically for multi-document QA. However, existing research has predominantly focused on addressing multi-document questions through a single LLM call, without explicitly distinguishing among individual documents within the query set. 

Crucially, analytical queries require comprehensive multi-step analysis across documents. While prior work has proposed frameworks for such multi-step, multi-document QA systems \citep{anderson2024design, shankar2024docetl}, standardized benchmarks for evaluating this capability remain publicly unavailable, and their documents are very short. To address this gap, we present MuDABench, a novel benchmark for Multi-Document Analysis and targeting scenarios involving document sets exceeding the context window of a single long-context LLM.

\section{Benchmark}

We collected 589 documents from US and Chinese listed companies with explicit metadata. Second, we set up about 5-38 PDF documents after each question, which is more than all existing work in terms of document pages and far exceeds the maximum LLM context. In the following, we will introduce our document types and annotation process in turn, and finally introduce our evaluation metrics.

\subsection{Document Source} 
Our document collection constitutes the most comprehensive repository of financial documents among available benchmarks. We systematically crawled annual reports, corporate announcements, and ESG report documents from two primary sources: cninfo\footnote{cninfo: https://www.cninfo.com.cn/} and SEC\footnote{SEC: https://www.sec.gov/}. 

\textbf{Annual reports} contain comprehensive disclosures of listed companies' operational status, published annually. These documents feature extensive structured tabular data.

\textbf{Announcements} represent ad-hoc disclosures by listed companies, with significant proportions of scanned documents.

\textbf{ESG reports} disclose corporate performance in environmental, social responsibility, and governance. These documents are characterized by complex visual layouts, including richly colored backgrounds and extensive pictorial elements \citep{li2025esg,zhang2025benchmarking,li2026deepread}.

\begin{figure}[t]
    \centering
    \includegraphics[width=1\linewidth]{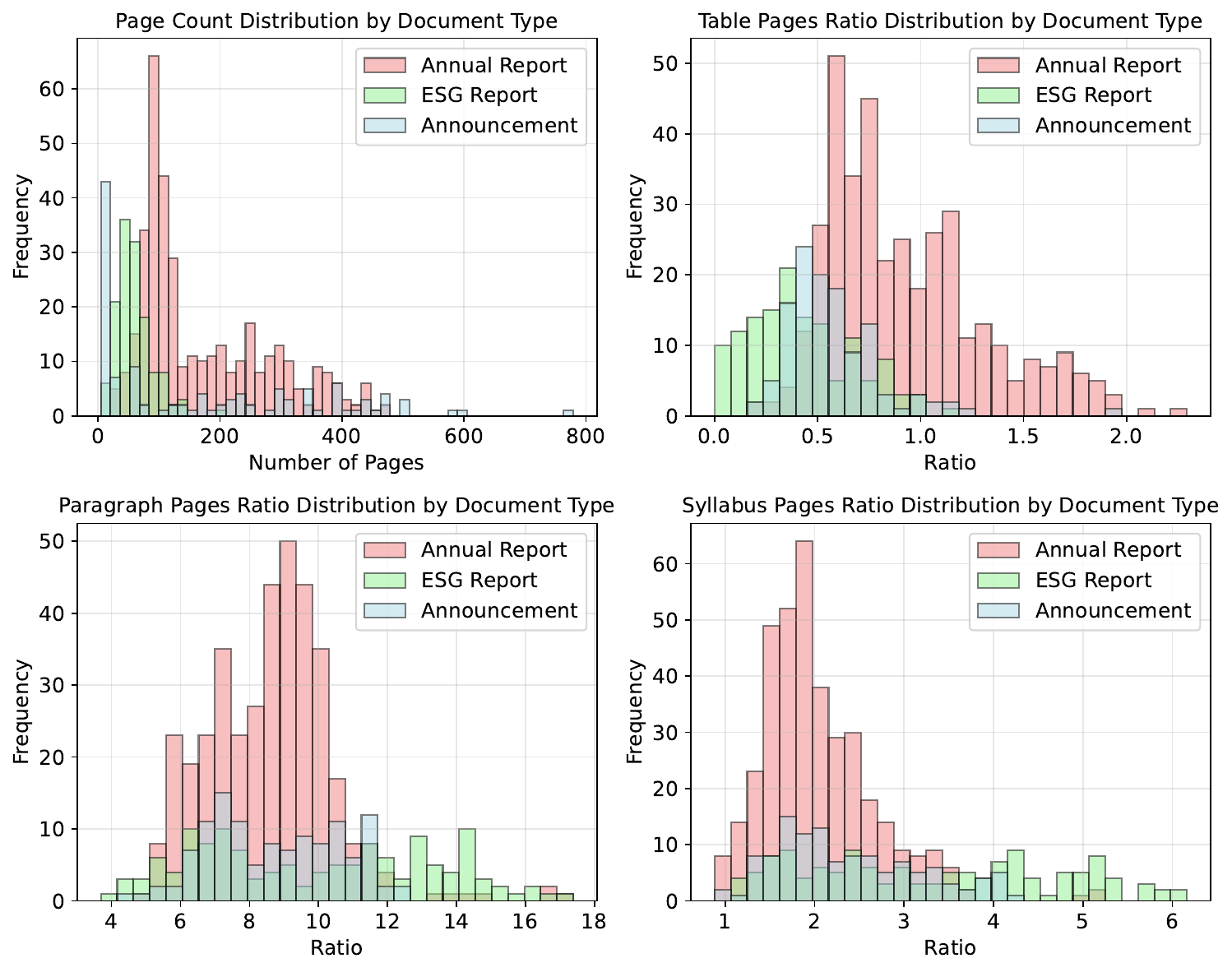}
    \caption{Document Elements Distribution. The distribution is divided into three sections: annual report, ESG report and announcement}
    \label{fig:pdf_label}
\end{figure}

This heterogeneous document format distribution ensures benchmark diversity and fits~\citet{hui2024uda}'s emphasis on the importance of parsing for document QA. Figure \ref{fig:pdf_label} illustrates the format distribution following parsing through an advanced commercial PDF processing tool provided by \citet{pdfparser2025}\footnote{ChatDOC: https://chatdoc.com/}.  

\subsection{Metadata Annotation}
Each document is accompanied by metadata, such as subject category and author's name in academic contexts, or date and coverage area in news domains. In our benchmark, each document is annotated with three metadata fields:

\textbf{Ticker symbol}: Identifies the company associated with the document.

\textbf{Fiscal year}: Indicates the period the document covers, distinct from its publication year.
 
\textbf{Document type}: Classified as annual reports (US or CN), ESG reports, meeting of shareholders announcement, or profit distribution equity announcements.

This metadata can be used to filter or prioritize documents without accessing their specific content.

\subsection{Question Annotation}
We employ a distant supervision annotation strategy~\cite{yang2018dcfee} combined with expert curation to construct our benchmark. First, we leverage authoritative financial databases to curate a comprehensive repository of structured data points, encompassing metrics such as revenue, executive details, dividends, and social responsibility indicators. These data are systematically organized into a master spreadsheet where each row corresponds to a document \( D_i \), indexed by metadata fields \( M_i \).

To generate the questions \( Q_j \), financial domain experts designed natural language question templates targeting specific analytical tasks (e.g., trend analysis, peer comparison, see more in Appendix~\ref{ap:sample}). These templates are instantiated using the structured data to produce diverse and realistic queries. Crucially, to facilitate robust evaluation via an LLM-as-a-judge, experts manually transcribed the specific structured indicators required for each question into natural language statements. These descriptive statements constitute the intermediate information set \( \mathcal{S}_j \), serving as the fine-grained ground truth for verifying whether the model has correctly extracted the necessary facts from the documents.

Formally, for each question \( Q_j \), we sample a set of \( k \) relevant documents to form the collection \( \mathcal{D}_j = \{ D_{j1}, D_{j2}, \ldots, D_{jk} \} \). The final dataset is formalized as:
\begin{equation}
\mathcal{X} = \left\{ \left( Q_j, \mathcal{D}_j, \mathcal{M}_j, \mathcal{S}_j \right) \mid j = 1, 2, \ldots, n \right\},
\end{equation}
where \( \mathcal{M}_j \) represents the metadata set for documents in \( \mathcal{D}_j \), \( \mathcal{S}_j \) denotes the set of natural language fact descriptions derived from the structured data, and \( n \) is the total number of samples.

\subsection{Question Grouping}
Roughly speaking, we categorize the benchmarks into \textbf{Simple} and \textbf{Complex} problems, based on three key dimensions: the volume of single-document information required, the complexity of numerical computation involved, and the depth of logical reasoning demanded to derive the answer.

For instance, a typical simple problem is formulated as: \textit{Please calculate the variance of the company's total cost in 2021 based on your knowledge base.} In contrast, a more complex version of the same problem would be: \textit{Please calculate the variance of total costs for companies audited by Big 4 accounting firms in your knowledge base for the year 2021.} The increased difficulty here is reflected in multiple layers of reasoning: first, identifying all companies in the dataset that meet the "audited by Big 4" criterion (which may require cross-referencing multiple documents or verifying implicit attributes); second, extracting total cost figures for only those filtered entities; and third, performing the variance calculation on this subset of data. Such a problem thus demands both conditional filtering of information and multi-step logical integration, distinguishing it from the straightforward extraction-computation pipeline of simple questions. More cases are in the Appendix~\ref{ap:sample}.

\subsection{Annotation Verification} 
Despite the existence of specialized databases as a reference, manual labeling may also contain errors. Therefore, we adopt a multi-document problem $Q$, use DeepSeek R1 to generate a single document query $q_i$ for every single document, and then input it into an RAG system, to get an answer $a_i$ on the document, and if there is any contradiction with $S_i$, then the problem is manually re-labeled or the question description needs to be modified. However, since ChatDOC is unable to adjust the number of recalled chunks, we did not include it in our subsequent experiments.

\subsection{Evaluation Metrics}
\label{subsec:evaluation-metrics}

We evaluate each system with three metrics: \textbf{process accuracy}, \textbf{final-answer accuracy}, and \textbf{full accuracy}. Among them, \textbf{final-answer accuracy} is our primary end-task metric, while \textbf{process accuracy} is mainly used as a diagnostic signal for intermediate extraction quality. We note that process coverage can be less reliable when equivalent evidence can be expressed in multiple non-atomic fact forms.

\noindent\textbf{Final-answer accuracy.}
For each question \(Q_i \in \mathcal{Q}\), let \(A_i\) be the gold final answer and \(\hat{A}_i\) be the model prediction.  
Let \(T_i \in \{0,1\}\) denote whether \(\hat{A}_i\) is semantically equivalent to \(A_i\) (judged by an LLM):
\begin{equation}
\text{Accuracy}_{\text{final}}
=
\frac{1}{|\mathcal{Q}|}\sum_i T_i.
\label{eq:acc_final}
\end{equation}

\noindent\textbf{Process accuracy.}
Let \(\mathcal{S}_i\) denote the gold set of minimal supporting facts for \(Q_i\), and \(\mathcal{I}_i\) denote the facts extracted by the system.

\textit{(a) Standard RAG (question-level).}
For standard RAG systems (single retrieved context, no explicit document alignment), we estimate fact coverage by judging how many gold supporting facts in \(\mathcal{S}_i\) are semantically supported by the extracted information \(\mathcal{I}_i\). Formally, we write
\begin{equation}
C_i=\frac{|\mathcal{I}_i \cap \mathcal{S}_i|}{|\mathcal{S}_i|},
\end{equation}
where the intersection denotes judge-determined semantic matches rather than exact string identity.
Because a single judge may overestimate coverage, we apply a double-check judge that estimates the error/missing ratio:
\begin{equation}
E_i=\frac{\#\text{ incorrect or missing facts in }\mathcal{S}_i}{|\mathcal{S}_i|}.
\end{equation}
We then use conservative coverage
\begin{equation}
\tilde{C}_i=\min\!\bigl(C_i,\;1-E_i\bigr).
\end{equation}
(Manual verification: agreement improves from \(16/30\) to \(26/30\).)

\textit{(b) Document-grounded workflow (cell-wise on aligned rows).}
Since MuDABench is derived from remotely annotated structured data, it is natural to evaluate how well the required table content can be reconstructed when answering such questions. We therefore evaluate process quality in a cell-wise manner on aligned rows. Let \(\mathcal{C}_i\) be the set of required gold metric cells across all aligned rows for question \(Q_i\), and \(\hat{\mathcal{C}}_i\) be the subset of correctly extracted cells:
\begin{equation}
C_i^{\text{cell}}=\frac{|\hat{\mathcal{C}}_i|}{|\mathcal{C}_i|}.
\end{equation}
To assess the reliability of this cell-wise judge, we manually audited 30 cases and compared the automatic cell-level decisions against human verification. The resulting cell-level agreement was \(81.93\%\).
To unify both settings, define the per-question process score as
\begin{equation}
P_i=
\begin{cases}
\tilde{C}_i, & \text{standard RAG},\\
C_i^{\text{cell}}, & \text{document-grounded workflow}.
\end{cases}
\end{equation}
Then process accuracy is defined as
\begin{equation}
\text{Accuracy}_{\text{process}}
=
\frac{1}{|\mathcal{Q}|}\sum_i P_i.
\label{eq:acc_process}
\end{equation}

\noindent\textbf{Full accuracy.}
Finally, we report a strict joint metric: a sample is counted as correct only if process is fully correct and final answer is correct.
Let \(m_i=\mathbf{1}[P_i=1]\). Then
\begin{equation}
\text{Accuracy}_{\text{full}}
=
\frac{1}{|\mathcal{Q}|}\sum_i m_i\,T_i.
\label{eq:acc_full}
\end{equation}

% This metric captures end-to-end reliability: a prediction is counted as correct only if all required intermediate facts are recovered and the final aggregated answer is also correct.
% 保持算法位置不变（内容已根据您之前的要求优化）

\section{Methodology}

% 保持图片位置不变
\begin{figure*}[t]
    \centering
    \includegraphics[width=0.8\linewidth]{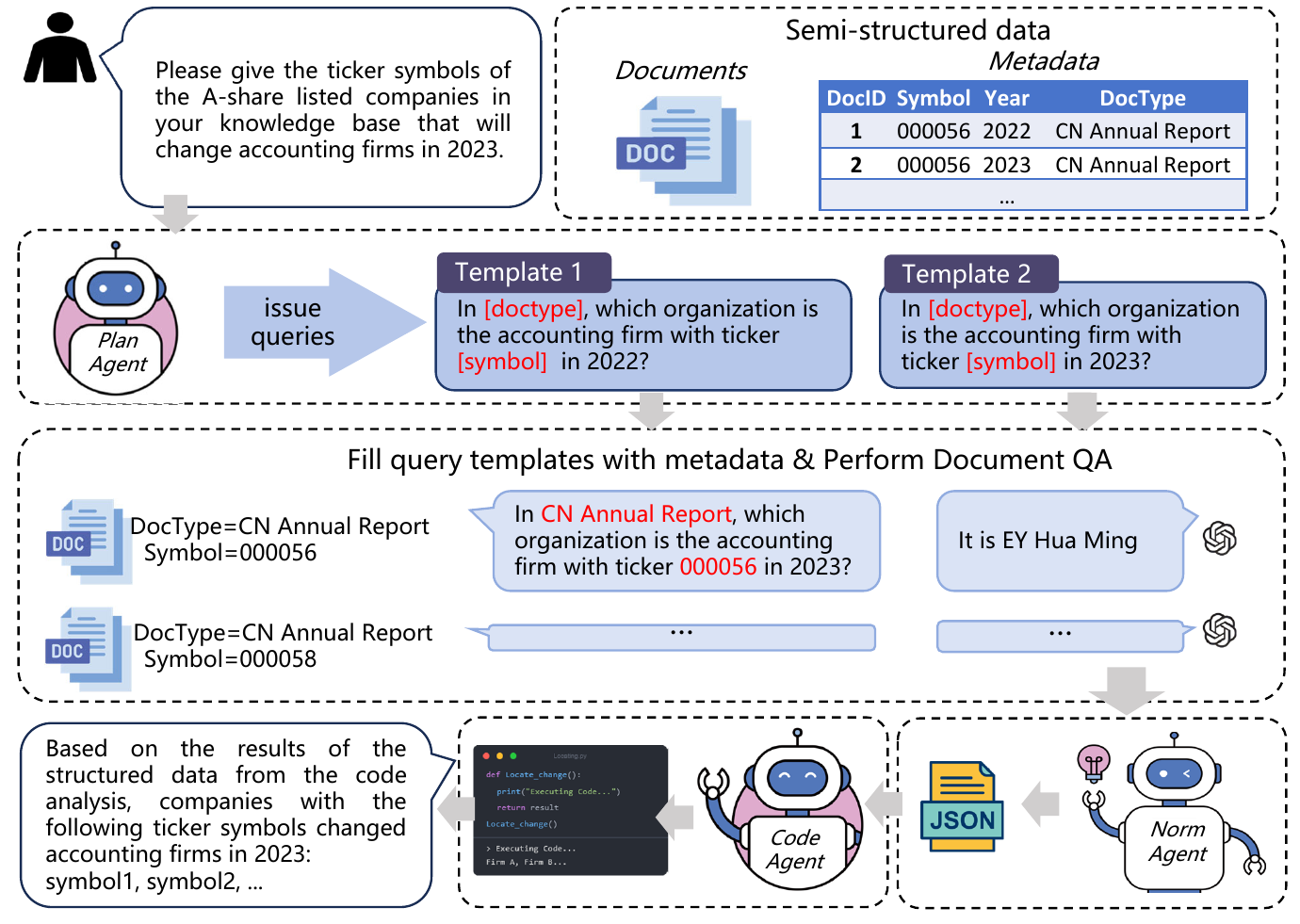}
    \caption{Document agentic workflow. A planning agent uses a metadata schema to generate sub-queries, an end-to-end RAG system answers single-document queries, then responses are normalized to JSON and analyzed by code to obtain the final answer.}
    \label{fig:agent}
\end{figure*}
\begin{algorithm}[t]
\caption{Metadata-Aware Multi-Agent Analytic QA Workflow}
\label{alg:agentic_workflow}
\small
\begin{algorithmic}[1]
\Require Query $Q$, document collection $\mathcal{D}=\{D_1,\dots,D_n\}$, metadata $\mathcal{M}=\{M_1,\dots,M_n\}$, batch size $B$
\Ensure Final answer $A$

\State \textcolor{gray}{\textit{// Phase 1: Planning}}
\State $\mathcal{T} \gets \Func{PlanAgent}(Q,\mathcal{M}_{\text{schema}})$ 
\Comment{Generate sub-query templates with optional metadata restrictions}

\State \textcolor{gray}{\textit{// Phase 2: Metadata-Guided Extraction}}
\State $\mathcal{Q}_{\text{pairs}} \gets \emptyset$
\For{$i = 1$ \textbf{to} $n$}
    \For{\textbf{each} $T_j \in \mathcal{T}$}
        \If{$\Func{SatisfyRestriction}(M_i,T_j)$}
            \State $q_{i,j} \gets \Func{FillTemplate}(T_j,M_i)$
            \Comment{Instantiate the template using document metadata}
            \State $a_{i,j} \gets \Func{RAGSystem}(D_i,q_{i,j})$
            \Comment{Single-document targeted extraction}
            \State $\mathcal{Q}_{\text{pairs}} \gets \mathcal{Q}_{\text{pairs}} \cup \{(M_i,q_{i,j},a_{i,j})\}$
        \EndIf
    \EndFor
\EndFor

\State \textcolor{gray}{\textit{// Phase 3: Schema Definition and Batch Normalization}}
\State $S_{\text{json}} \gets \Func{DefineSchema}(\Func{Sample}(\mathcal{Q}_{\text{pairs}}), Q)$
\State $\mathcal{J} \gets \emptyset$
\State $K \gets \left\lceil |\mathcal{Q}_{\text{pairs}}| / B \right\rceil$
\For{$k = 1$ \textbf{to} $K$}
    \State $\mathcal{B}_k \gets \Func{GetBatch}(\mathcal{Q}_{\text{pairs}},k,B)$
    \State $\mathcal{J}_{k} \gets \Func{NormAgent}(\mathcal{B}_k,S_{\text{json}})$
    \State $\mathcal{J} \gets \mathcal{J} \cup \mathcal{J}_{k}$
\EndFor

\State \textcolor{gray}{\textit{// Phase 4: Programmatic Analysis}}
\State $p_{\mathcal{J}} \gets \Func{SaveJSON}(\mathcal{J})$
\Comment{Save full structured records to an external file}
\State $\mathcal{J}_{\text{demo}} \gets \Func{Sample}(\mathcal{J})$
\Comment{Provide only examples to the code agent}
\State $C_{\text{code}} \gets \Func{CodeAgent}(Q,\mathcal{J}_{\text{demo}},S_{\text{json}},p_{\mathcal{J}})$
\State $R_{\text{exec}} \gets \Func{Execute}(C_{\text{code}},p_{\mathcal{J}})$

\State \textcolor{gray}{\textit{// Phase 5: Final Synthesis}}
\State $A \gets \Func{FinalAgent}(Q,R_{\text{exec}},\mathcal{J}_{\text{demo}})$
\State \Return $A$

\end{algorithmic}
\end{algorithm}
MuDABench presents a unique challenge where document collections exceed the context window of current LLMs, rendering single-pass ingestion infeasible~\citep{huang2023advancing,levy2025more}. To address this, we propose a scalable \textbf{Multi-Agent Analytic QA Workflow} that explicitly orchestrates multi-step reasoning over large-scale repositories. A key feature of this approach is its ability to scale to processing hundreds or thousands of documents. The procedure is detailed in Algorithm~\ref{alg:agentic_workflow} and visualized in Figure~\ref{fig:agent}. The workflow consists of four specialized components:

\textbf{Scalable Planning Agent:} Instead of retrieving documents immediately, this agent decomposes the global query $Q$ into question templates to be asked on each document. The template can be filled with metadata of documents. This abstraction minimizes planning errors and ensures the approach scales to collections of arbitrary size.

\textbf{Document-Level Information Extractor:} We perform targeted extraction by instantiating the query templates for each document $D_i$ using its specific metadata $M_i$. A standard document RAG system then processes these instantiated queries in parallel, producing intermediate textual evidence that captures local facts.
    
\textbf{Scalable Norm Agent:} To enable downstream programmatic reasoning, this agent converts unstructured extraction transcripts into structured JSON records. Crucially, to avoid context overflow when processing thousands of documents, we adopt a \emph{batch-iterative} strategy: a schema is defined from a small sample, and subsequent records are normalized in batches under this unified schema.
    
\textbf{Scalable Code Agent:} Rather than feeding the entire extracted information into the LLM, we provide the agent with the schema and some examples. The agent synthesizes a program to perform analysis over the full structured dataset $\mathcal{J}$~\citep{wang2024executable}, yielding the final answer $A$.

\begin{table*}[t]
\centering

% \resizebox{\textwidth}{!}{%
\small
\begin{tabular}{lcccccc}
\toprule
\rowcolor{headergray}
 & \multicolumn{3}{c}{\textbf{Simple}} & \multicolumn{3}{c}{\textbf{Complex}} \\
\cmidrule(lr){2-4} \cmidrule(lr){5-7}
\rowcolor{headergray}
\textbf{Model} & \textbf{$Acc_{process}$} & \textbf{$Acc_{final}$} & \textbf{$Acc_{full}$} & \textbf{$Acc_{process}$} & \textbf{$Acc_{final}$} & \textbf{$Acc_{full}$} \\
\midrule
GPT 4o + Chunk $ = 1\,|\mathcal{D}|$   & 0.1572 & 0.0663 & 0.0241 & 0.1459 & \textbf{0.0482} & 0.0181 \\
GPT 4o + Chunk $ = 1.5\,|\mathcal{D}|$ & 0.1761 & 0.0964 & 0.0301 & 0.1801 & \textbf{0.0482} & \textbf{0.0241} \\
GPT 4o + Chunk $ = 2\,|\mathcal{D}|$   & 0.1793 & \textbf{0.1265} & \textbf{0.0422} & 0.2212 & 0.0361 & 0.0181 \\
GPT 4o + Chunk $ = 2.5\,|\mathcal{D}|$ & \textbf{0.2163} & 0.1084 & 0.0301 & \textbf{0.2623} & \textbf{0.0482} & 0.0120 \\
\midrule
GPT 4o + Chunk $ = 1\,|\mathcal{D}|$ + Metadata   & 0.1338 & 0.1084 & 0.0422 & 0.1398 & 0.0301 & \textbf{0.0181} \\
GPT 4o + Chunk $ = 1.5\,|\mathcal{D}|$ + Metadata & 0.1620 & 0.1145 & 0.0301 & 0.1773 & 0.0181 & 0.0120 \\
GPT 4o + Chunk $ = 2\,|\mathcal{D}|$ + Metadata   & 0.1978 & \textbf{0.1386} & 0.0422 & 0.2232 & 0.0361 & \textbf{0.0181} \\
GPT 4o + Chunk $ = 2.5\,|\mathcal{D}|$ + Metadata & \textbf{0.2514} & 0.1325 & \textbf{0.0542} & \textbf{0.2522} & \textbf{0.0422} & 0.0120 \\
\midrule
WF w/ GPT 4o + Chunk $ = 1$             & 0.4179 & 0.0667 & 0.0000 & 0.4021 & 0.0667 & 0.0095 \\
WF w/ GPT 4.1 mini + Chunk $ = 3$       & 0.5803 & \textbf{0.2430} & 0.0654 & 0.5338 & 0.0865 & 0.0673 \\
WF w/ GPT 4.1 mini + Chunk $ = 5$       & 0.5888 & 0.2243 & \textbf{0.0748} & \textbf{0.5749} & \textbf{0.1619} & \textbf{0.1143} \\
Noise WF w/ GPT 4.1 mini + Chunk $ = 5$ & \textbf{0.5961} & 0.1636 & 0.0727 & 0.5680 & 0.1238 & 0.0762 \\

\midrule
\rowcolor{humanblue}
Human Performance & 0.8940 & 0.8334 & 0.7334 & 0.8120 & 0.7334 & 0.6667 \\
\bottomrule
\end{tabular}
\caption{Experiment Results. Bolded labels indicate the optimal setup within each model group. The $Acc_{process}$ metric has been adjusted to reflect the full evaluation set.}
\label{tab:result}
\end{table*}

\begin{table*}[!t]
\centering

\small
\begin{tabular}{lc|cc|cc|cc}
\toprule
\rowcolor{headergray}
\multirow{1}{*}{\textbf{Document Category}} & \multirow{1}{*}{\textbf{Avg. Tokens / Doc}} & \multicolumn{2}{c|}{\textbf{Chunk = 1}} & \multicolumn{2}{c|}{\textbf{Chunk = 3}} & \multicolumn{2}{c}{\textbf{Chunk = 5}} \\
\rowcolor{headergray}
 &  & \textbf{Simple} & \textbf{Complex} & \textbf{Simple} & \textbf{Complex} & \textbf{Simple} & \textbf{Complex} \\
\midrule
A-share Annual Report (CN)  & 499k & 0.4696 & 0.4555 & 0.6537 & 0.6674 & 0.6447 & 0.6570 \\
A-share ESG Report (CN)     & 72k  & 0.3998 & 0.3898 & 0.6067 & 0.4813 & 0.5865 & 0.4992 \\
A-share Announcement (CN)   & 144k & 0.3903 & 0.3786 & 0.5222 & 0.4976 & 0.5542 & 0.5575 \\
US Stock Annual Report (EN) & 120k & 0.4472 & 0.3955 & 0.3167 & 0.5374 & 0.4643 & 0.7375 \\
\bottomrule
\end{tabular}

\caption{Impact of Chunk Number on $Acc_{process}$ across Document Categories, including average document length.}
\label{tab:chunk_impact}
\end{table*}

\vspace{-0.2cm}
\section{Experiment}

\subsection{Experimental Setup}

The experiments are conducted on MuDABench. As a natural baseline, we employ a RAG system~\citep{lewis2020retrieval} over the multi-document corpus. We consider two prompt variants: one that omits document metadata and one that injects all metadata into the prompt (detailed in Appendix~\ref{sub:all_prompt}). Both use OpenAI's File Search as the retrieval layer with \texttt{GPT-4o-2024-11-20} as the reader~\citep{openai_gpt4o_2024_11_20}. To study the effect of recall, we set the number of retrieved chunks to $|\mathcal{D}|$ and then increase it to $1.5 \times |\mathcal{D}|$, $2 \times |\mathcal{D}|$, and $2.5 \times |\mathcal{D}|$. All other hyperparameters follow OpenAI defaults.

We also evaluate our proposed agentic workflow. 
We use \texttt{DeepSeek-R1-0528}~\citep{guo2025deepseek} for the planning and code agent, \texttt{DeepSeek-Chat-V3-0324}~\citep{liu2024deepseek} for the normalization agent, and OpenAI file search for single document QA.
% The planning agent, based on \texttt{deepseek-r1-0528}~\citep{guo2025deepseek}, generates sub-query templates; GPT-4o with RAG executes these templates; the normalization agent, powered by \texttt{deepseek-chat-v3-0324}~\citep{liu2024deepseek}, converts transcripts to JSON; and \texttt{deepseek-r1-0528} performs code generation and produces final answers. 
To control cost, we use \texttt{gpt-4o-2024-11-20} for one high-budget chunk of the workflow (approximately 30{,}000 tokens per document) and \texttt{gpt-4.1-mini-2025-04-14}~\citep{openai_gpt41mini} for the remaining workflow experiments. 
To study robustness under noisy contexts, we additionally inject $0.5 \times |\mathcal{D}|$ irrelevant documents in the 5-chunk workflow setting, e.g., adding 2023 filings to questions about 2021--2022.

Given the task complexity, we adopt an LLM-as-judge protocol \citep{gu2024survey}. Evaluation has two components: (1) assessing information extraction to obtain \(C_i\), and (2) verifying answer correctness, from which we derive the three metrics defined in the previous Section \ref{subsec:evaluation-metrics}. For RAG, since it lacks explicit intermediate outputs, we evaluate the retrieval recall by checking if the retrieved chunks cover the gold facts $\mathcal{S}_i$. Specific prompts are provided in Appendix~\ref{sub:all_prompt}. All LLMs are run with temperature \(0\) for reproducibility. Two volunteers answer a subset of the benchmark to estimate human performance.

\begin{table}[t]
    \centering
    \small

    \begin{tabular}{lccc}
        \toprule
        \rowcolor{headergray}
        \textbf{Metric} & \textbf{Simple} & \textbf{Complex} & \textbf{Avg}\\
        \midrule
        \multicolumn{4}{l}{\textit{Independent accuracy}}\\
        Planning       & 86.7\% & 93.3\% & 90.0\%\\
        Extraction     & 40.0\% & 20.0\% & 30.0\%\\
        Normalization  & 100.0\%& 100.0\%& 100.0\%\\
        Code           & 93.3\% & 93.3\% & 93.3\%\\
        \midrule
        \multicolumn{4}{l}{\textit{Accuracy given all previous steps correct}}\\
        Planning       & 86.7\% & 93.3\% & 90.0\%\\
        Extraction     & 38.5\% & 14.3\% & 25.9\%\\
        Normalization  & 100.0\%& 100.0\%& 100.0\%\\
        Code           & 80.0\% & 100.0\%& 85.7\%\\
        \bottomrule
    \end{tabular}

        \caption{The percentage of correct in each step}
    \label{tab:error_performance}
\end{table}

\subsection{Main Result}
The experimental results, summarized in Table~\ref{tab:result}, highlight the significant challenges posed by MuDABench and the distinct behaviors of different system architectures.

\textbf{Standard RAG pipelines struggle with multi-document aggregation, and metadata injection only provides limited gains.} 
Even when powered by GPT-4o, commercial RAG systems exhibit clear structural limitations on MuDABench. Increasing the number of retrieved chunks generally improves evidence coverage, as reflected by higher process accuracy, but this gain does not translate reliably into better final answers: on simple questions, final-answer accuracy improves only up to a point and then fluctuates, while on complex questions it remains consistently low despite better coverage. This pattern suggests that the main bottleneck is not merely retrieval recall, but the model's ability to synthesize fragmented evidence into a correct aggregated conclusion. Explicitly incorporating document metadata offers only partial mitigation. Although metadata can provide a coarse global structure and sometimes improves performance, the overall gains remain limited, indicating that neither larger retrieval budgets nor metadata cues are sufficient without a more structured reasoning workflow.

\textbf{Agentic workflows significantly improve end-to-end answer quality.}
The proposed agentic workflow substantially outperforms direct RAG in final-answer accuracy, demonstrating the value of decomposing extraction and reasoning into modular stages. Although the workflow also tends to achieve higher process-coverage scores, we treat this metric as a diagnostic signal rather than the primary basis for comparison, because equivalent evidence can be represented in flexible and sometimes non-atomic ways. Under stronger chunk budgets, the workflow delivers markedly better end-to-end performance than standard RAG. Furthermore, robustness analysis reveals that injecting noise (irrelevant documents) causes a noticeable drop in final-answer accuracy, especially on complex questions, indicating that downstream aggregation and reasoning remain important bottlenecks.

\begin{figure}[h]
    \centering
    \includegraphics[width=1\linewidth]{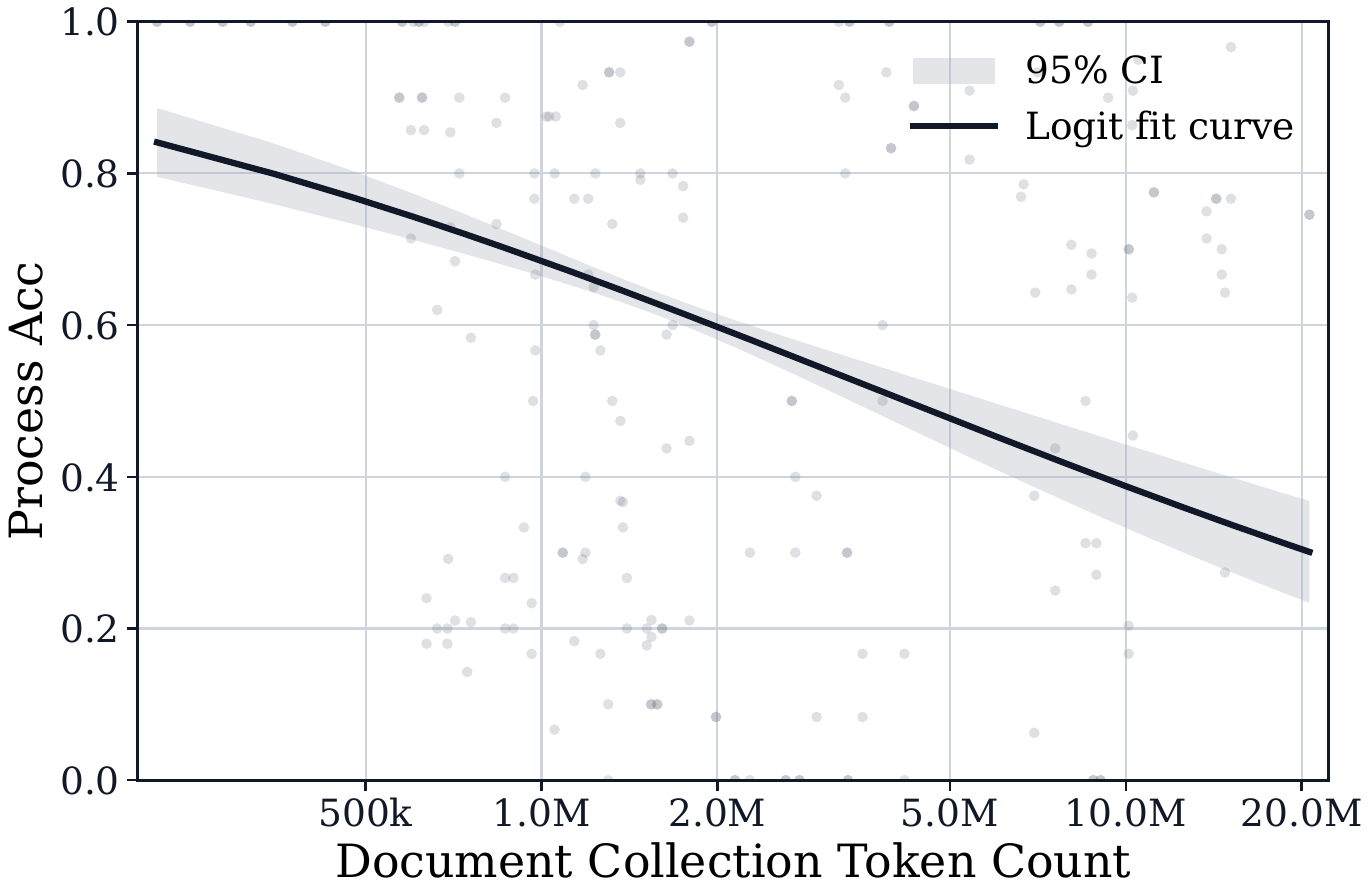}
    \caption{The Impact of Document Collection Token Count on Document Information Extraction}
    \label{fig:logit}
\end{figure}
\subsection{Fine-Grained Error Analysis}
We conduct a fine-grained diagnostic study on 30 randomly selected examples under the 5-chunk workflow setting, with results summarized in Table~\ref{tab:error_performance}. Since final-answer accuracy is our primary metric, we use process-oriented analysis mainly to identify bottlenecks rather than as a fully reliable standalone measure. Process coverage can be noisy because equivalent evidence may be expressed in non-atomic forms: for example, ``an increase of 6\% from 2021 to 2022'' may correspond to two separate facts such as ``100k in 2021'' and ``106k in 2022'', while the final answer may only require the growth rate. With this caveat, the results still indicate that document-level information extraction is the main bottleneck. We also find planning errors caused by insufficient financial domain knowledge (Appendix~\ref{case_plan}), and coding errors mainly due to encoding or JSON path-reading issues.

To further explore why document-level extraction remains difficult, we estimated a logit model with fixed question type effects. The resulting curve, shown in Figure~\ref{fig:logit}, suggests that information extraction becomes more difficult as document length increases.

This length-dependent performance degradation is further corroborated by the category-wise breakdown in Table~\ref{tab:chunk_impact}. In general, shorter document categories tend to be easier to process, although the pattern is not uniform across all settings. For example, A-share ESG reports, which have a relatively small average token count, achieve competitive extraction accuracy, but they are not consistently the best-performing category under every chunk budget; in several settings, A-share annual reports or U.S. annual reports perform better. Overall, Table~\ref{tab:chunk_impact} suggests that document length is an important factor, but extraction difficulty also depends on document type and language.

\section{Conclusion}

We present MuDABench, the first large-scale benchmark designed for complex, metadata-driven analysis over semi-structured document collections. It requires navigating over 80,000 pages of financial documents. Evaluations reveal that standard RAG systems struggle at this scale. While our metadata-aware agentic workflow significantly improves final-answer accuracy over the baselines, it still trails human performance. We hope MuDABench serves as a rigorous testbed for future scalable document analysis systems.

\section*{Limitations}
Our work is restricted to the financial domain due to the scarcity of dense semi-structured data elsewhere. We also limited the dataset size. Although distant supervision can increase the size easily, the testing is costly and does not yield new discoveries. Finally, because financial atomic facts involve numerous issues related to granularity and equivalence, they must be handled with care when evaluating different QA systems.

\section*{Ethical Considerations}
The documents in MuDABench are collected from publicly available financial disclosures, ensuring that no private or non-public personal information is compromised. While we utilizes real-world financial figures, it is intended solely for the research.  We used AI for minor language polishing. To ensure accuracy, we work with the community to correct any potential annotation errors in the dataset on an ongoing basis; therefore, the evaluation results in this paper may not be up to date. 

\section*{Acknowledgments}
This work has been supported by the National Natural Science Foundation of China (No. 62206265, 62076231).

\bibliography{acl}

\clearpage
\onecolumn
\appendix
\section{Appendix}
\subsection{Comparison on the Same Selected Subset}
\label{ap:comp}

To provide a fair comparison between human performance and agentic workflows under the same evaluation scope, we report results on the same selected subset used in our human perfermance. This subset is divided into \textbf{Simple} and \textbf{Complex} cases. Table~\ref{tab:wf_selected_cases} shows that, although stronger workflow configurations improve both process and final-answer performance on this subset, all agentic systems still remain substantially below human performance.

\begin{table}[h]
\centering
\small
\caption{Performance comparison on the subset of Human Performance.}
\begin{tabular}{lccc|ccc}
\toprule
\textbf{Model} & \multicolumn{3}{c|}{\textbf{Simple}} & \multicolumn{3}{c}{\textbf{Complex}} \\
\cmidrule(lr){2-4}\cmidrule(lr){5-7}
& \textbf{$Acc_{process}$} & \textbf{$Acc_{final}$} & \textbf{$Acc_{full}$}
& \textbf{$Acc_{process}$} & \textbf{$Acc_{final}$} & \textbf{$Acc_{full}$} \\
\midrule
WF w/ GPT 4o + Chunk $ = 1$             & 0.5936 & 0.1429 & 0.0000 & 0.5396 & \textbf{0.3333} & 0.0667 \\
WF w/ GPT 4.1 mini + Chunk $ = 3$       & 0.6983 & \textbf{0.2667} & \textbf{0.2000} & 0.6978 & \textbf{0.3333} & \textbf{0.2667} \\
WF w/ GPT 4.1 mini + Chunk $ = 5$       & 0.6897 & 0.2000 & \textbf{0.2000} & \textbf{0.7230} & \textbf{0.3333} & 0.2000 \\
Noise WF w/ GPT 4.1 mini + Chunk $ = 5$ & \textbf{0.7328} & \textbf{0.2667} & 0.1333 & 0.6906 & 0.2667 & 0.1333 \\
\midrule
\rowcolor{humanblue}
Human Performance & 0.8940 & 0.8334 & 0.7334 & 0.8120 & 0.7334 & 0.6667 \\
\bottomrule
\end{tabular}

\label{tab:wf_selected_cases}
\end{table}
\subsection{Case Study}
\subsubsection{Planning Errors Prior to Information Extraction}
\label{case_plan}
In complex financial question answering, a substantial fraction of failures arise already in the planning phase, before any document-level extraction is performed. These errors are typically rooted in insufficient domain knowledge about market conventions and disclosure practices, which leads the agent to design sub-queries that are structurally misaligned with the underlying task.

\begin{figure*}[t]
    \centering
    \includegraphics[width=1\linewidth]{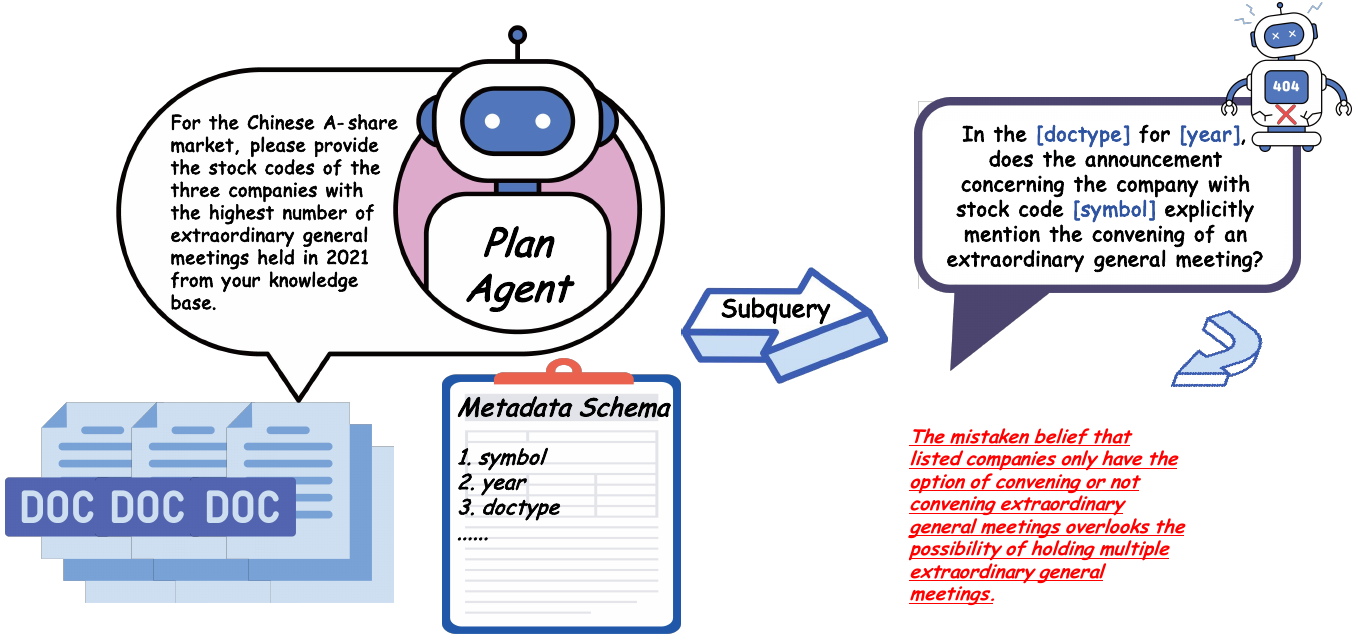}
    \caption{Case Study: Planning errors stemming from a lack of knowledge regarding shareholder meetings}
    \label{fig:meeting-error}
\end{figure*}

Figure~\ref{fig:meeting-error} illustrates a representative planning error driven by an incorrect mental model of corporate event frequencies. The user query explicitly requests the companies with the \textit{highest number} of extraordinary general meetings. However, the Plan Agent generates a sub-query that only verifies the \textit{existence} of such a meeting (i.e., whether at least one was convened). This effectively reduces a counting problem to a binary classification problem, and ignores the fact that listed firms may hold multiple extraordinary general meetings within a single fiscal year. As a result, the downstream pipeline never triggers the aggregation logic necessary to rank companies by meeting counts.

\begin{figure*}[t]
    \centering
    \includegraphics[width=1\linewidth]{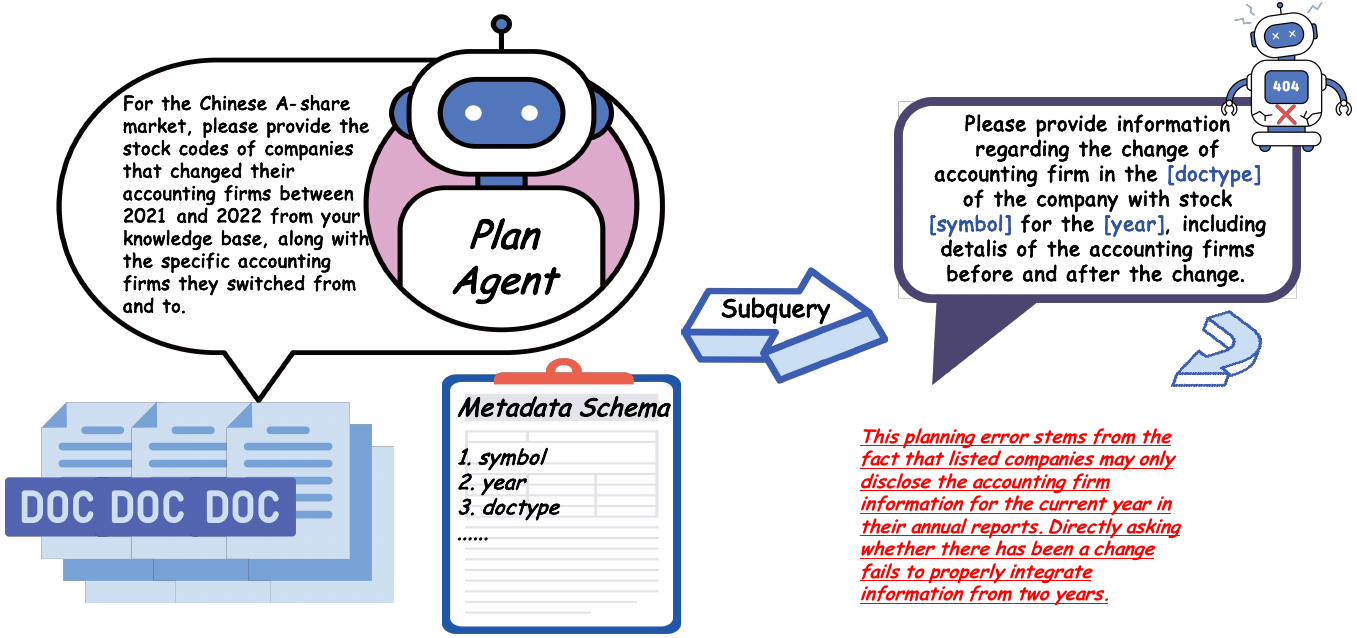}
    \caption{Case Study: Planning errors stemming from a lack of established annual report preparation protocols}
    \label{fig:protocol-error}
\end{figure*}

Figure~\ref{fig:protocol-error} shows a second class of planning failure related to annual report disclosure protocols. The task requires identifying changes in accounting firms between 2021 and 2022. Instead of decomposing the task into two extraction steps—retrieving the engaged accounting firm in 2021 and in 2022, and then comparing them—the agent attempts to locate an explicit textual description of the ``change'' event within a single document. This strategy contradicts standard reporting practices, where annual reports usually disclose only the currently engaged firm for that specific fiscal year rather than the full transition history. Because the plan does not incorporate this protocol knowledge, the system fails to construct the necessary multi-hop reasoning chain and the retrieval stage subsequently breaks down.

\subsubsection{Errors After Information Extraction}
\begin{figure*}[t]
    \centering
    \includegraphics[width=1\linewidth]{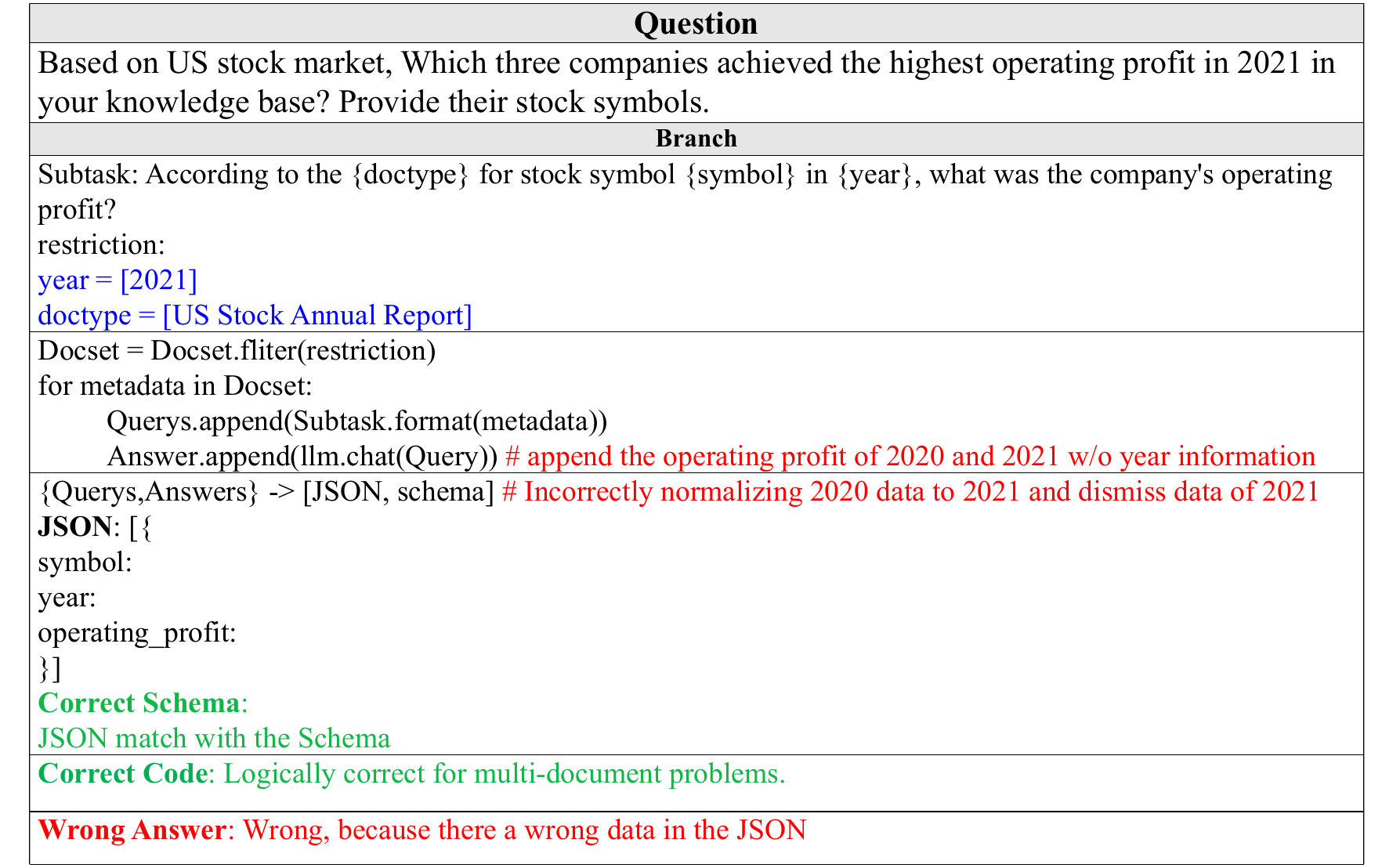}
    \caption{Case Study: Ambiguous Information Extraction and Normalization Failure}
    \label{fig:norm-case}
\end{figure*}

Even when the planning stage is successful, errors can still emerge in the information extraction and normalization stages. Figure~\ref{fig:norm-case} illustrates a failure caused by coupled extraction and normalization issues. Financial reports routinely present data for both the current and previous fiscal years within the same table or paragraph. Models often struggle to disambiguate which subset of these values corresponds to the target reporting period, leading to extractions that conflate multi-year information. When such ambiguous entries are later merged with strictly single-year values from other documents, the resulting heterogeneity in temporal scope severely complicates normalization and downstream analysis.

\begin{figure*}[t]
    \centering
    \includegraphics[width=1\linewidth]{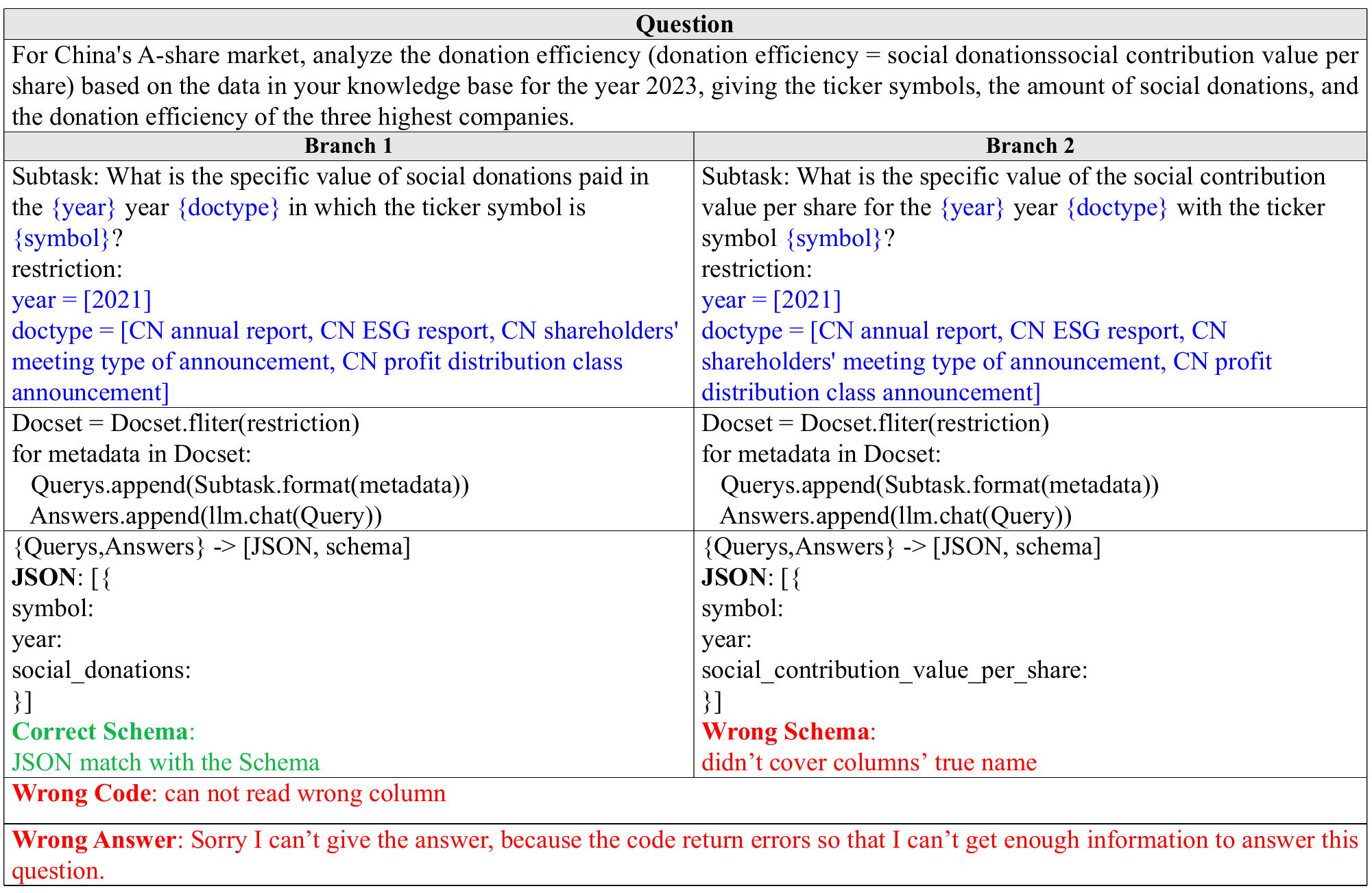}
    \caption{Case Study: Schema Alignment Error in Code Execution}
    \label{fig:code-case}
\end{figure*}

Figure~\ref{fig:code-case} depicts a related failure mode at the schema alignment stage. Here, the extracted JSON records deviate from the predefined schema, for example by introducing inconsistent field names or missing mandatory keys. Although these deviations may appear minor at the textual level, they cause runtime exceptions in the code analysis agent and prevent the execution of otherwise valid analytical programs. This highlights that robust large-scale multi-document analysis requires not only accurate extraction but also strict adherence to a stable schema across all stages of the workflow.

\subsection{Data Source Details}

To construct our dataset, we sourced data from several authoritative financial databases widely used by researchers. These sources are detailed in Table~\ref{tab:data_sourece}.

\begin{table}[ht]
\centering
\ResetTable{
\begin{tabular}{lcc}
\toprule
\textbf{Data} & \textbf{Source} & \textbf{Description} \\
\midrule
CN Annual Report & https://www.cninfo.com.cn/ & Annual financial disclosures of Chinese listed companies \\
US Annual Report & https://www.sec.gov/ & Annual financial disclosures (e.g., 10-K) of US listed companies \\
CN ESG Report & https://www.cninfo.com.cn/ & Environmental, Social, and Governance reports of Chinese companies \\
CN Announcement & https://www.wind.com.cn/ & Ad-hoc corporate announcements of Chinese listed companies \\
CN Financial Data & https://data.csmar.com/ & Structured financial metrics and market data for China \\
US Financial Data & https://data.csmar.com/ & Structured financial metrics and market data for the US \\
\bottomrule
\end{tabular}}
\caption{Data Source. Among them, cninfo and SEC are free public data, and wind and CSMAR are commercial databases.}
\label{tab:data_sourece}
\end{table}

\subsection{Benchmark Construction Detail}

Overall, our annotation is semi-automated by means of distant supervision. First we download some structured data in \texttt{.csv} format from CSMAR, which stores financial and non-financial metrics of listed companies, and then we match each document with the data in this \texttt{.csv}, and then our annotators just need to convert each metric into natural language descriptions and define a series of questions, through the permutations and combinations our annotation strategy can be easily scaled to generate large amounts of data.

\subsubsection{Announcement of merger strategy}

For the announcement category, since there is no standardization of announcements made by different companies, but it can be confirmed that the disclosure of information has a lag, so we use a MERGE strategy, that is, the same category of announcements of 2021 and 2022 as the announcement of the fiscal year 2021. This ensures that the information in the document includes all the information of the fiscal year 2021.

\subsubsection{Cross-year Problem}

It's worth mentioning that for the cross-year problem, we generate an intermediate information collection $\mathcal{S}$ that is half the size of the document collection $\mathcal{D}$, and we merge two years of information for the same company. An intuitive example would be, \textit{Which are the three companies with the highest revenue growth rates from 2022--2023 among U.S. publicly traded companies in your knowledge base}, and the $S_i$ for this question would look like this: \textit{Company A's revenue in 2022 was xx, and in 2023 it was xx, an increase of xx}, and in this way we encourage the QA system to do some simple computational reasoning in the information extraction. This design doesn't affect the design of the evaluation metrics in our paper.

\subsection{Model Details}

In this paper we use a range of expensive commercial modeling services, from pdf parsing and model measurement, as shown in Table~\ref{tab:model_sourece}.

\begin{table*}[t]
\centering
\ResetTable{
\begin{tabular}{lcc}
\toprule
\textbf{Model} & \textbf{Source} & \textbf{Usage} \\
\midrule
ChatDOC & https://chatdoc.com/ & PDF parsing \\
OpenAI GPT 4o & https://platform.openai.com/docs/models/gpt-4o?snapshot=gpt-4o-2024-11-20 & LLM for RAG \\
OpenAI GPT 4.1 mini & https://platform.openai.com/docs/models/gpt-4.1-mini & LLM for RAG \\
OpenAI File Search & https://platform.openai.com/docs/guides/tools-file-search & Retrieve \\
DeepSeek R1 & https://openrouter.ai/deepseek/deepseek-r1-0528 & Plan and Code Agent \\
DeepSeek V3 & https://openrouter.ai/deepseek/deepseek-chat-v3-0324 & Norm Agent \\
DeepSeek V3.2 & https://openrouter.ai/deepseek/deepseek-v3.2 & Cell-wise Rejudge Judge \\
Kimi k2 & https://openrouter.ai/moonshotai/kimi-k2 & Other Judge Tasks \\
\bottomrule
\end{tabular}}
\caption{Model Source. All of the above except DeepSeek and Kimi are closed-source models, and all of them call commercial API services in the experiments of this paper.}
\label{tab:model_sourece}
\end{table*}

\subsection{Judge Details}

We use different judge models for different evaluation components. In particular, \texttt{DeepSeek-V3.2} is used for the cell-wise rejudge setting on aligned rows, which requires fine-grained field-level matching between extracted evidence and aligned gold rows, while \texttt{Kimi K2} is used for the remaining correctness judgments. All judge models are run at temperature 0.
\subsection{Prompt Template}
\label{sub:all_prompt}
\subsubsection{Prompt Template of adding metadata to RAG}

\begin{promptbox}[title={Prompt: adding metadata to RAG}]
The list of document metadata you can query is as follows: 
\textbf{[document\_info]}

You need to answer the question:
\textbf{[question]}
\end{promptbox}

\subsubsection{Prompt Template of Plan Agent}

\begin{promptbox}[title={Prompt: Plan Agent}]
\begin{itemize}
  \item \textbf{Initial Prompt}: You are a multi-document problem decomposition and query assistant. The user is provided with a metadata description of a complex task and multiple documents. Each document has a unique metadata identifier, and instead of answering the final question directly, you need to design specialized query templates for each document so that you can extract information from it that is relevant to the overall task and organize these subqueries in a reasonable order. Also you need to generate the query templates based on using the same language as the language of user's task. For example, if the user speaks Chinese, you generate the query template in Chinese.

  \item \textbf{Question template specification}
  \begin{enumerate}
    \item \textbf{Single Document Oriented}: Each sub-question template must focus on a single document to ensure that the required information can be located within that document.
    \item \textbf{Semantic Mutual Exclusivity}: different templates should be independent of each other in meaning and not duplicated; populated to be able to ask questions naturally and smoothly and logically self-consistent.
    \item \textbf{Metadata Placement}: all available metadata fields are placed by \{\} in at least one template, while all metadata fields must use naming consistent with the metadata field description when placed in templates.
    \item \textbf{Metadata constraints (optional)}: for each sub-question template, metadata can be constrained. If you think that answering a multi-document question requires that this sub-question template restricts a certain (some) metadata, please use the name of that metadata as a label, with all possible values listed in list format within the label. For example: \texttt{"restriction": \{"year": ["2021", "2022"]\}}. If there is no restriction, the field can be left out.
  \end{enumerate}

  \item \textbf{User-supplied information}:
  \begin{itemize}
    \item Multi-document task description: \textbf{[task]}
    \item Available metadata fields: \textbf{[metadata\_description]}
  \end{itemize}

  \item \textbf{Output format} (JSON only; no extra text)
\end{itemize}
{\footnotesize\ttfamily
\begin{tabbing}
\hspace{1.5em}\=\hspace{1.5em}\=\hspace{1.5em}\=\kill
[\\
\>\{\\
\>\>"subtask": "Format template for subquestion",\\
\>\>"restriction": \{"year": ["2021","2022"]\}\\
\>\},\\
\>\{\\
\>\>"subtask": "Format template for subquestion"\\
\>\}\\
]
\end{tabbing}
}
\end{promptbox}

\subsubsection{Prompt Template of Norm Agent (Stage 1)}

\begin{promptbox}[title={Prompt: Norm Agent (schema)}]
\begin{itemize}
\item \textbf{Initial Prompt}:
You are an expert in structured data extraction, extract structured data related to complex tasks directly from multiple sub-question answered conversations and transform it into json format, the final output json data should be easily scalable when transform similar conversations to json. Also you need to generate schema based on using the same language as the users's task language. For example, if the user speaks Chinese, you generate the query template in Chinese.
\item \textbf{Input information}:
\begin{enumerate}
  \item Complex problem description: \textbf{[task]}
  \item Complete dialog record of sub-question answers: \textbf{[multi\_conversation]}
\end{enumerate}
\item \textbf{Processing requirements}:
\begin{itemize}
  \item extract all specific data values (numbers, options, measurements, etc.) related to the complex task and standardize the units of measurement.
  \item identify variable types and add metadata:
  \begin{itemize}
    \item \textbf{Classification variables}: list the values that actually occur
    \item \textbf{Ordinal variables}: Preserve sequential relationships
    \item \textbf{Quantitative variables}: specify the units used
  \end{itemize}
  \item For information not provided by the user, keep it as null.
  \item Naming of variables should reflect the meaning of the variable and the unit of measure.
  \item ideal json structure must just have one level, no nested structure and can be easily analysis by python code.
\end{itemize}
\item \textbf{Output format}:
Please output data surrounded by \texttt{<json>}...\texttt{</json>} (must be a JSON \texttt{list[dict]}), and add a brief schema with \texttt{<des>}...\texttt{</des>}.
\end{itemize}
\end{promptbox}

\subsubsection{Prompt Template of Norm Agent (Stage 2)}

\begin{promptbox}[title={Prompt: Norm Agent (continuation)}]
\begin{itemize}
  \item \textbf{Initial Prompt}:
  You are a json continuation helper that converts new conversation records to json data based on the original conversation record converted json data (\texttt{list[dict]}) format provided by the user, which is easy to merge with the original data.

  \item \textbf{Input information}:
  \begin{enumerate}
    \item json data of the original conversation record: \textbf{[json]}
    \item new conversation record: \textbf{[new\_conversation]}
  \end{enumerate}

  \item \textbf{Processing requirements}:
  \begin{enumerate}
    \item Convert the new conversation record to json format, making sure it is consistent with the original data structure.
    \item Maintain consistency in variable naming and units of measure.
    \item Make sure the new json data can be seamlessly connected to the original data.
  \end{enumerate}

  \item \textbf{Output Format}:
  Please enclose your extended section with \texttt{<json>}...\texttt{</json>} (a \texttt{list[dict]} that can be concatenated to the original data with \texttt{+} in Python).
\end{itemize}
\end{promptbox}

\subsubsection{Prompt Template of Code Agent}

\begin{promptbox}[title={Prompt: Code Agent}]
\begin{itemize}
  \item \textbf{Initial Prompt}:
  You are a question answering expert, the user will provide a complex task and multiple copies of related json data and their paths, you need to write code based on this data to get the data needed to answer the question. Please note for each available, the user only provides the few shot of original json data (\texttt{list[dict]}) for you to write code for this task, and the user will provide the path to the json data, you need to read the json data from the user-provided path, execute the code will directly solve the task ideally.

  \item \textbf{Input information}:
  \begin{enumerate}
    \item task description: \textbf{[task]}
    \item available json data (few shot): \textbf{[json\_data]}
    \item path to the json data: \textbf{[json\_path]}
    \item schema of the json data: \textbf{[json\_schema]}
  \end{enumerate}

  \item \textbf{Processing requirements}
  \begin{enumerate}
    \item Analyze the task description to identify key issues and data requirements.
    \item Based on the json data provided, write executable python code to read the json data from the user-provided path and extract the required information.
    \item The code output should be readable, ideally the code output should answer the task directly.
  \end{enumerate}

  \item \textbf{Output format}:
  Wrap your code in \texttt{<execute>}...\texttt{</execute>} and you can add necessary explanations outside the tags.
\end{itemize}
\end{promptbox}

\subsubsection{Prompt of Final Answer}

\begin{promptbox}[title={Prompt: Final Answer}]
The user is provided with a complex task, JSON data description, Python code, and run results. Produce the final concise answer (in the user's language).

\textbf{Inputs}: \textbf{[task]}, \textbf{[data]}, \textbf{[code]}, \textbf{[code\_resp]}.
\end{promptbox}

\subsubsection{Prompt Template of judging Information Extraction (cell-wise on aligned rows)}

\begin{promptbox}[title={Prompt: Judge --- Information Extraction}]
\textbf{The user will provide:}
\begin{enumerate}
  \item Multi-document question: \textbf{[question]}
  \item Source table headers: \textbf{[source\_headers]}
  \item One gold source row (\texttt{row\_index = \{row\_index\}}): \textbf{[source\_row]}
  \item Aligned target document metadata (already matched by the dataset, do not judge metadata again): \textbf{[aligned\_doc\_meta]}
  \item Metric columns to judge from this row (\texttt{metric\_total = \{metric\_total\}}): \textbf{[metric\_columns]}
  \item Complete dialog record of all sub-question answers on this document: \textbf{[agent\_conversation]}
\end{enumerate}

\textbf{Evaluation Criteria:}
\begin{enumerate}
  \item Judge only against this single gold row and this single document dialog.
  \item This gold row has already been aligned to the correct document by the dataset metadata. Treat the document identity as given.
  \item Do not score metadata fields in this step. Judge only the metric columns in this row one by one.
  \item A metric column counts as correct only if the document dialog contains the same core fact under the correct symbol and year context.
  \item For numeric columns, treat the extraction as correct if the integer digits and the first decimal place are correct, unless the task clearly requires exact discrete identifiers.
  \item Extra information in the dialog does not hurt correctness.
  \item Count each column at most once.
\end{enumerate}

TASK:
\begin{enumerate}
  \item Return only the metric fields that are correctly supported by the dialog.
\end{enumerate}

\textbf{Constraints:}
\begin{enumerate}
  \item \texttt{correct\_metric\_fields} must contain only field names from the provided metric columns.
  \item If none are correct, return an empty list.
\end{enumerate}

\textbf{Return JSON only:}
{\footnotesize\ttfamily
\begin{tabbing}
\hspace{1.5em}\=\hspace{1.5em}\=\kill
\{\\
\>"correct\_metric\_fields": ["<metric field name>"]\\
\}
\end{tabbing}
}
\end{promptbox}

\subsubsection{Prompt Template of Judging RAG Information Extraction (The correct side)}

\begin{promptbox}[title={Prompt: Judge --- RAG Information Extraction (The correct side)}]
\textbf{The user will provide:}
\begin{enumerate}
  \item Information (\texttt{list}) needed to answer the question (total\_required = \{\texttt{len\_source}\}): \textbf{[source\_answer]}
  \item Information extracted by the model (each part separated by \texttt{<next chunk>} is independent of each other): \textbf{[info]}
\end{enumerate}

\textbf{Evaluation Criteria:}
\begin{enumerate}
  \item If the information extracted by the model contains the key information with true entity from the reference, the correct extraction is added by one.
  \item If the information does not match the entity or does not provide the entity information, it is judged incorrect and the number of correct extractions remains unchanged.
  \item If the key information is missing or does not match the reference, it is judged incorrect and the number of correct extractions remains unchanged.
  \item If the extraction is missing or does not match the reference information, it is judged incorrect and the number of correct extractions remains unchanged.
  \item If the extracted information involves numerical values, the model is considered correct as long as it correctly extracts integer digits and the first decimal place.
\end{enumerate}

Note: If the extraction of the model contains information other than the reference information or uses a different language, this does not affect the determination. Multiple correct extractions of the same information are counted only once.

TASK: Calculate the number of intersections between the information extracted by the model and the information that needs to be extracted.

\textbf{Return JSON:}

{\footnotesize\ttfamily
\begin{tabbing}
\hspace{1.5em}\=\hspace{1.5em}\=\kill
\{\\
\>"correct\_extractions": <number of correct entries>,\\
\>"total\_required": len\_source,\\
\>"explanation": "..."\\
\}
\end{tabbing}
}
\end{promptbox}

\subsubsection{Prompt Template of Judging RAG Information Extraction (The incorrect side)}
\begin{promptbox}[title={Prompt: Judge --- RAG Information Extraction (The incorrect side)}]
\textbf{The user will provide:}
\textbf{The user will provide:}
\begin{enumerate}
  \item Information (\texttt{list}) needed to answer the question (total\_required = \{\texttt{len\_source}\}): \textbf{[source\_answer]}
  \item Information extracted by the model (each part separated by \texttt{<next chunk>} is independent of each other): \textbf{[info]}
\end{enumerate}
\textbf{Evaluation Criteria: }

\begin{enumerate}
    \item Treat the reference information list as the gold standard.
    \item For each required entry in the reference list, check all extracted chunks:
    \item  If at least one chunk correctly contains the key information with the true entity, this entry is counted as correctly extracted.
    \item  If the key information is missing, conflicts with the reference (wrong entity/value), or is otherwise incorrect, this entry is counted as an error.
    \item  If the extracted information involves numerical values, the model is considered correct as long as it correctly extracts integer digits and the first decimal place.
    \item  Extra information that is not in the reference list, or uses a different language, does not affect the judgment. Only the correctness of the required entries is considered.
    \item  Multiple correct extractions of the same reference entry are counted only once.
    \item  error\_extractions is the number of required entries that are incorrect or missing (i.e., entries in the reference list that do not have a correct extraction).
\end{enumerate}
TASK:You are asked to consider each part of the model output separated by <next chunk> as a piece of information extracted by the model, and compute:
\begin{enumerate}
    \item error\_extractions = number of incorrect or missing entries among the required information.
    \item total\_required = {len\_source}.
\end{enumerate}
\textbf{Return JSON only:}
{\footnotesize\ttfamily
\begin{tabbing}
\hspace{1.5em}\=\hspace{1.5em}\=\kill
\{\\
\>"error\_extractions": <number of incorrect or missing entries>,\\
\>"total\_required": \{len\_source\},\\
\>"explanation": "your simple explanation"\\
\}
\end{tabbing}
}
\end{promptbox}

\subsubsection{Prompt Template of Judging Final Answer}

\begin{promptbox}[title={Prompt: Judge — Final Answer}]
\textbf{The user will provide:}
\begin{enumerate}
  \item Multi-document question: \textbf{[question]}
  \item Multi-document reference answer: \textbf{[final\_answer]}
  \item Model's answer: \textbf{[model\_answer]}
\end{enumerate}
\textbf{Evaluation Criteria: }
\begin{enumerate}
  \item Model's answer will be judged as correct if it contains the key information in the reference answer.
  \item If the key information is missing or the answer does not match the reference answer, the answer will be judged as incorrect.
  \item If the answer is missing or does not match the reference answer, the answer is judged as incorrect.
  \item If the answer involves a numerical value, it is considered correct as long as the model answers the whole number of digits and the first decimal place correctly.
\end{enumerate}
Note: If a model answer contains additional information beyond the reference answer or use different language, this does not affect the judgment as correct.

TASK: Determine whether the model's final answer is correct with respect to the reference answer.
\textbf{Return JSON}:

{\footnotesize\ttfamily
\begin{tabbing}
\hspace{1.5em}\=\hspace{1.5em}\=\kill
\{"is\_correct": true/false, "explanation": "..."\}
\end{tabbing}
}
\end{promptbox}

\subsection{Benchmark Examples}
\label{ap:sample}

\begin{table}[h]
\centering
\small
\setlength{\tabcolsep}{5pt}
\renewcommand{\arraystretch}{1.3}
\caption{Some Examples of Our Benchmark}
\label{tab:sample-table}
\begin{tabularx}{\linewidth}{>{\raggedright\arraybackslash}X >{\raggedright\arraybackslash}X}
\toprule
\multicolumn{2}{c}{\textbf{Simple}} \\
\midrule
\textbf{Question} & \textbf{Source Answer} \\
\midrule
\textit{For China's A-share market, please provide the stock codes of the three companies with the highest capital adequacy ratio in 2021 from your knowledge base.}
&
Stock code [\text{symbol}] had a capital adequacy ratio of xx in [\text{year}]. \\

\textit{Please calculate the range of total operating costs for A-share companies in 2021 from your knowledge base.}
&
Based on [\text{year}] data, the maximum total operating cost for A-share companies was xx, the minimum was xx, and the range was xx. \\

\textit{Based on the US stock market, which three companies had the highest proportion of other business revenue in 2021 in your knowledge base? List their stock symbols.}
&
Company [\text{symbol}] had main business revenue of \$xx million and other business revenue of \$xx million in [\text{year}], with other revenue representing xx\% of total revenue. \\

\textit{Based on the US stock market, which three companies had the highest proportion of other business revenue in 2022 in your knowledge base? List their stock symbols.}
&
Company [\text{symbol}] had main business revenue of \$xx million and other business revenue of \$xx million in [\text{year}], with other revenue representing xx\% of total revenue. \\

\textit{For China's A-share market, which companies in your knowledge base had their equity registration date in 2023 advanced by more than two weeks compared to 2022? Provide their stock codes.}
&
Stock code [\text{symbol}] had an equity registration date of xx in [\text{year}] and xx in [\text{year+1}]. \\

\midrule
\multicolumn{2}{c}{\textbf{Complex}} \\
\midrule
\textbf{Question} & \textbf{Source Answer} \\
\midrule
\textit{Based on the US stock market, report the top three companies by total assets in 2022, including their stock symbols and exact asset values.}
&
Stock code [\text{symbol}] had total assets of \$xx in [\text{year}]. \\

\textit{For China's A-share market, please calculate the amount of change in the total cost of doing business in your knowledge base from 2021 to 2023 in terms of extreme variance.}
&
Based on data from [\text{year1}] to [\text{year2}], the maximum change in total operating cost for A-share companies was xx, the minimum change was xx, and the range of changes was xx. \\

\textit{Please provide the stock codes of the three companies in your knowledge base with the highest pre-tax dividend per share growth rates in the 2022--2023 annual distribution and their specific growth rates.}
&
Stock code [\text{symbol}] from \text{year} to \text{year+1}, the pre-tax dividend per share changes from xx to xx, with a growth rate of xx. \\

\textit{Based on the data in your knowledge base for the 2023 annual distribution, analyze the intervals between the share registration date and the ex-dividend date, and provide the ticker symbols of the three companies with the shortest intervals and the number of days between them.}
&
Stock code [\text{symbol}] in [\text{year}] yearly distribution had a share registration date of xx, ex-rights and ex-dividend date of xx, with an interval of xx days. \\

\textit{Please provide the stock codes of the outlier firms whose rate of change in capital adequacy, total operating costs, or net income exceeds the knowledge base mean $\pm$ twice the standard deviation over the period 2022--2023.}
&
Stock code [\text{symbol}] ratio of change in capital adequacy xx (knowledge base mean xx, standard deviation xx), ratio of change in total operating costs xx (knowledge base mean xx, standard deviation xx), and ratio of change in net income xx (knowledge base mean xx, standard deviation xx) over the period from [\text{year}] to [\text{year+1}]. \\

\bottomrule
\end{tabularx}
\end{table}
\end{document}